\documentclass{article}
\usepackage{spconf,amsmath,graphicx,hyperref}
\usepackage{comment}
\usepackage{subcaption, booktabs}
\usepackage{amssymb}

\title{EfficientIML: Efficient High-Resolution Image Manipulation Localization}
\name{Jinhan Li$^{1}$, Haoyang He$^{2}$, Lei Xie$^{2}$, Jiangning Zhang$^{2}$}
\address{$^{1}$ New York University, New York, NY, United States \quad $^{2}$ Zhejiang University, Hangzhou, China}

\begin{document}
%\ninept
%
\maketitle
\begin{abstract}
With imaging devices delivering ever-higher resolutions and the emerging diffusion-based forgery methods, current detectors trained only on traditional datasets (with splicing, copy-moving and object removal forgeries) lack exposure to this new manipulation type. To address this, we propose a novel high-resolution SIF dataset of 1,200+ diffusion-generated manipulations with semantically extracted masks. However, this also imposes a challenge on existing methods, as they face significant computational resource constraints due to their prohibitive computational complexities. Therefore, we propose a novel EfficientIML model with a lightweight, three-stage EfficientRWKV backbone. EfficientRWKV’s hybrid state-space and attention network captures global context and local details in parallel, while a multi-scale supervision strategy enforces consistency across hierarchical predictions. Extensive evaluations on our dataset and standard benchmarks demonstrate that our approach outperforms ViT–based and other SOTA lightweight baselines in localization performance, FLOPs and inference speed, underscoring its suitability for real-time forensic applications.
\end{abstract}
\section{Introduction}
\label{sec:intro}

Modern forensic analysis increasingly relies on Image Manipulation Localization (IML) to detect and precisely mask manipulated regions in digital images. With the development of high‐fidelity diffusion‐based inpainting methods such as Stable Diffusion\cite{rombach2021highresolution} and the proliferation of ultra–high–resolution imaging devices, traditional localization techniques must now handle subtler manipulations over vastly larger canvases.

Numerous benchmark datasets have driven progress in image forgery localization, including CASIA V1/V2\cite{Dong2013}\cite{pham2019hybrid}, Columbia\cite{hsu06crfcheck}, NIST16\cite{518026}, DEFACTO\cite{DEFACTODataset}, Coverage\cite{wen2016}, IMD\cite{Novozamsky_2020_WACV} and more recently, CIMD\cite{Zhang_2024}. Despite their importance, these collections are largely composed of low/medium-resolution images and simple artifact-based manipulations, which constrain their utility for training detectors aimed at identifying semantic, diffusion-generated forgeries at the high resolutions that are common in current applications. Although diffusion-based datasets such as CSI-IMD\cite{10943879} have been proposed, their low resolution (below 600×800) prevents alignment with the operational scales of modern detectors.

\begin{figure}[t]
\centering
\includegraphics[width=0.8\linewidth]{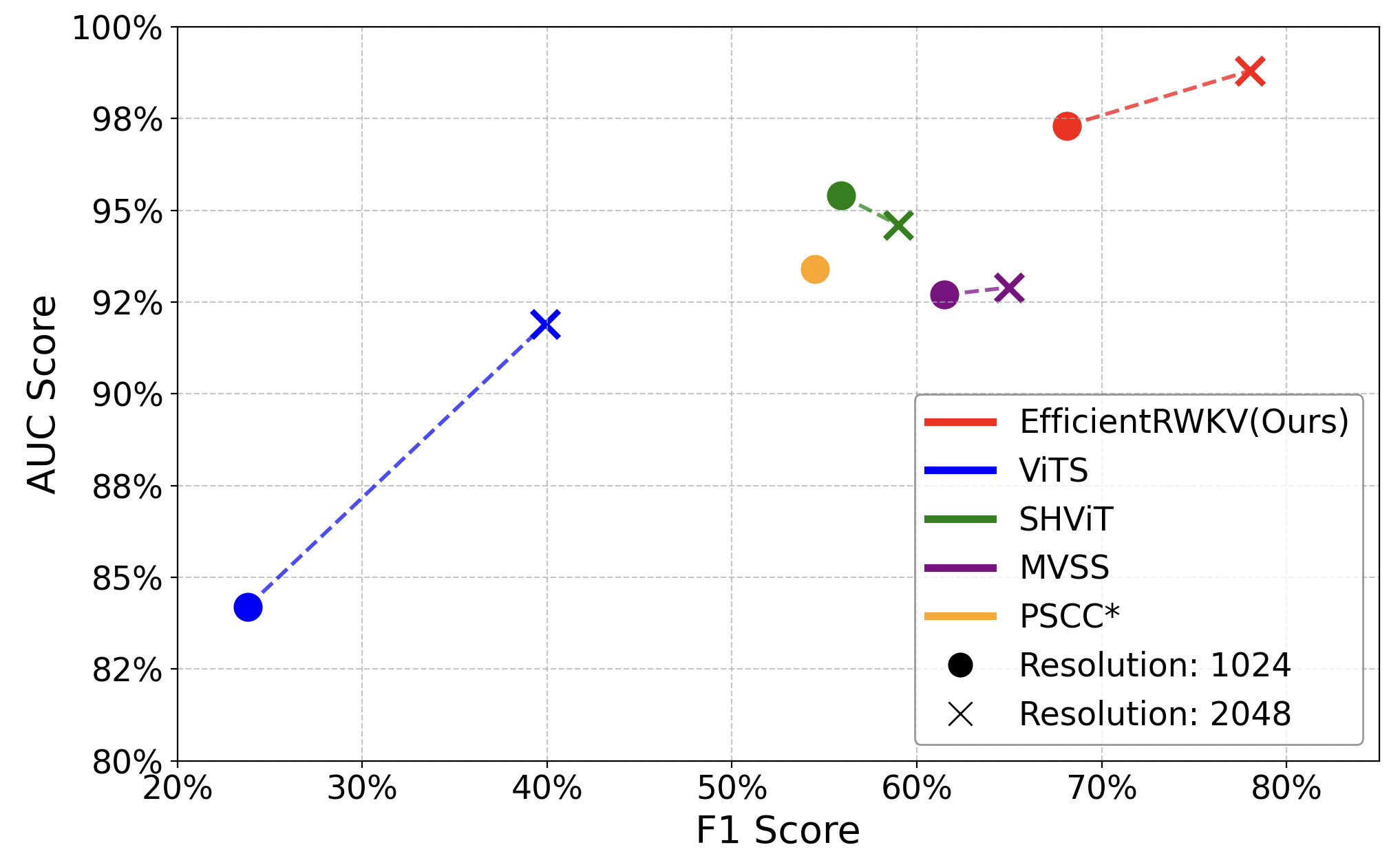}
\caption{\textbf{Performance of SOTA Architectures on SIF dataset}}
\label{graph_performace}
\vspace{-0.7em}
\end{figure}

To overcome these dataset gaps, we introduce the SIF (Semantic Inpainting Forgery) dataset, the first publicly released high–resolution semantic dataset exceeding 1024×1024. SIF comprises 1200+ carefully curated, diffusion‐generated inpainted images at up to 1800×1200 resolution, all hand-picked and human-validated by our group members and volunteers for semantic correctness and visual quality.

\begin{figure}[b]
\centering
\vspace{-1em}
\includegraphics[width=\linewidth]{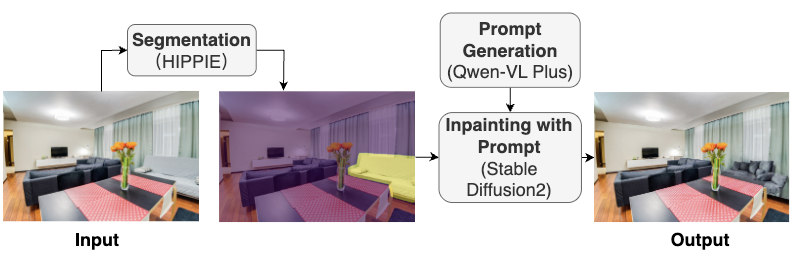}
\caption{\textbf{Pipeline for Creating SIF Dataset}}
\label{graph_dataset_create}
\vspace{-0.3em}
\end{figure}

\begin{figure*}[ht]
\centering
\includegraphics[width=0.85\linewidth]{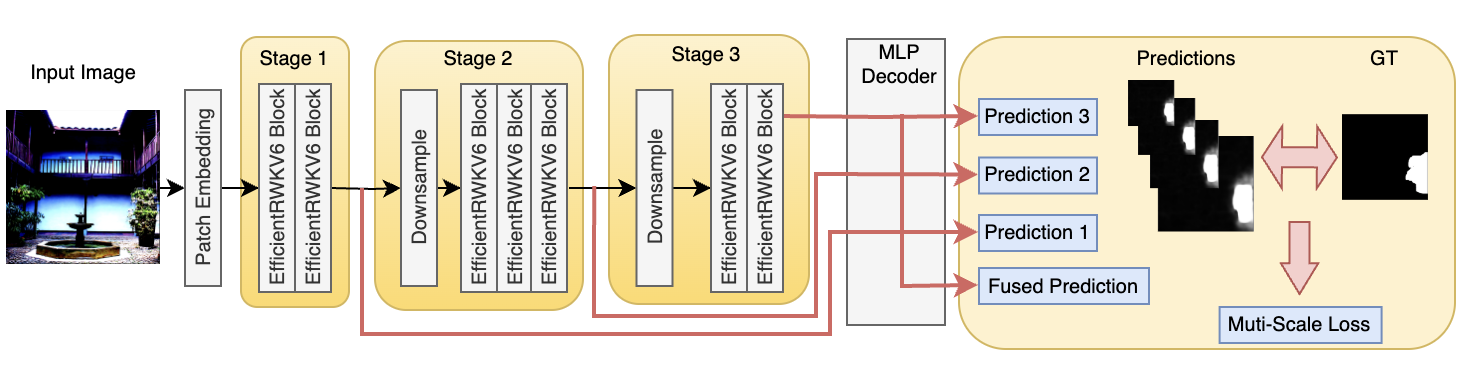}
\caption{\textbf{Macro Design}}
\label{graph_macro}
\vspace{-0.5em}
\end{figure*}

At the same time, localizing subtle forgeries in large images remains a formidable challenge. Vision transformer (ViT)–based detectors (such as IML-ViT\cite{ma2023imlvit}) suffer from quadratic complexity, making them impractical for on–device or real-time analysis. On the other hand, many CNN models such as MVSS-Net\cite{MVSS_2022TPAMI} either sacrifice long–range context or incur similar prohibitive FLOPs when scaled to higher resolutions. Existing lightweight architectures such as SHViT\cite{yun2024shvit} still demand over 380 GFLOPs at 2048×2048 inputs.

In this work, we address these limitations by proposing a novel EfficientRWKV backbone based on Vision–RWKV\cite{duan2024vrwkv}. It only requires a small fraction of the computational resources compared to the lightweight SOTA models, allowing it to operate on higher resolution datasets with limited computational resources. Even when input size grows to 2048×2048, we kept the computational cost under 100 GFLOPs. The primary contributions of this paper are:

\noindent1. \textbf{SIF dataset}, the first ultra-high–resolution semantic inpainting forgery dataset beyond 1024×1024.

\noindent2. A lightweight, three–stage \textbf{EfficientRWKV} backbone that fuses global state–space attention and local convolutional processing, with linear computational complexities.

\noindent3. A multi-scale decoder equipped with tailored loss weights.

\vspace{-0.5em}
\section{High-Resolution SIF Dataset}
\label{sec_dataset}

In this study, we introduce a high-resolution manipulated image dataset consisting of 1,228 carefully curated samples. Our construction pipeline is a modified version based on the Inpaint Anything framework\cite{yu2023inpaint} and the source images are selected from the SAM9K dataset\cite{chen2024sharegpt4v}. The procedure for dataset generation is illustrated in Fig.~\ref{graph_dataset_create}.

For an input image, we leverage the Hierarchical Open-vocabulary Universal Image Segmentation\cite{wang2023hierarchical} (HIPPIE) to segment for the object mask, and use a vision-language model, Qwen-VL Plus\cite{Qwen-VL}, to generate an appropriate prompt for inpainting. Then, the prompt, original image, and mask are input into Stable Diffusion 2~\cite{rombach2021highresolution}. This model replaces the masked region in the image with semantically coherent and visually consistent content, guided by the prompt. The result is a manipulated image that maintains high visual fidelity while incorporating subtle or significant alterations.

Approximately 5,000 manipulated images were generated through the above pipeline. From this pool, a final set of 1,228 images was manually selected based on visual realism, semantic coherence, and diversity of manipulations.

\section{Methodology: EfficientIML}
In this work, we propose a novel and efficient image manipulation localization framework EfficientIML (Fig.~\ref{graph_macro}) by integrating an optimized EfficientRWKV backbone with a customized multi-scale prediction scheme. Our pipeline builds upon the baseline structure of IML-ViT\cite{ma2023imlvit}, but introduces these two significant architectural innovations to improve both performance and efficiency.

\subsection{EfficientRWKV Backbone} 
We develop a lightweight attention backbone called \textbf{EfficientRWKV}, based on Vision RWKV6\cite{duan2024vrwkv}, made up of multiple EfficientRWKV blocks. It incorporates multi-scale embedding, stage-wise attention, and adaptive downsampling. 

\noindent\textbf{EfficientRWKV Block.} This block is a lightweight yet expressive building unit designed to capture both global and local visual dependencies while maintaining efficiency. Each EfficientRWKV Block consists of a sequential application of the following components, including depthwise convolution residual modules, a token mixer implemented via the EfficientRWKV Module, and Point-wise Feed Forward Networks (FFNs). Among them, the EfficientRWKV Module (Fig.~\ref{graph_micro}) performs multi-branch attention. It splits the input tensor $x$ along the channel dimension into three specialized branches:

\begin{figure}[t]
\centering
\includegraphics[width=\linewidth]{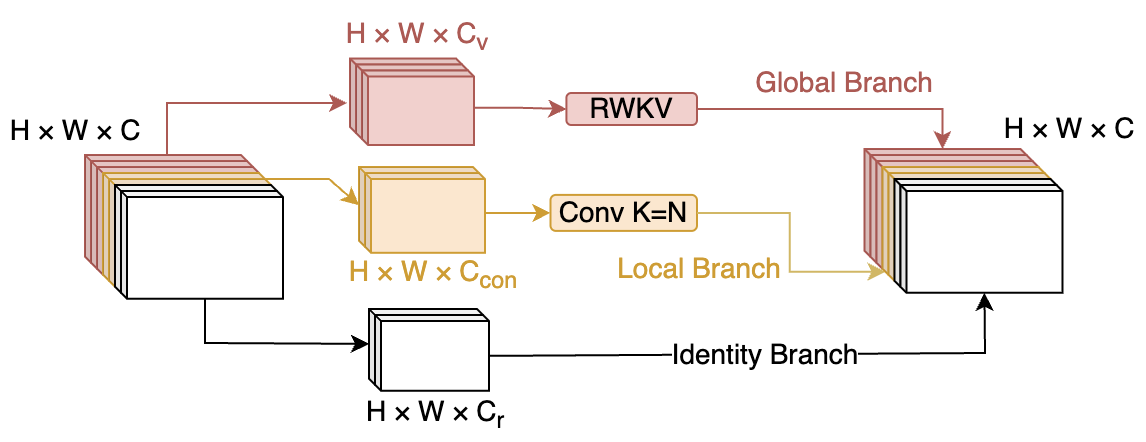}
\caption{\textbf{EfficientRWKV6 Module}}
\label{graph_micro}
\end{figure}

\begin{table*}[t]
  \centering
  \scriptsize 
  \captionsetup{justification=centering}
  \caption{\textbf{CASIAV2-training with resolution 1024×1024}}
  \setlength{\tabcolsep}{3pt}
  \vspace{-0.5em}
  % Row 1: F1 and AUC
  \begin{subtable}[t]{0.45\textwidth}
    \centering
    \caption{Pixel‐level F1 Score (\%)}
    \begin{tabular}{lccccccc}
    \toprule
     & CASIAV1 & Columbia & Coverage & NIST16 & IMD & CIMD & Average \\
    \midrule
    ViT-S\cite{ma2023imlvit}  &  32.7 & 24.1 &  4.8 & 15.7 & 19.3 & 9.8 &17.7\\
    SHViT\cite{yun2024shvit}  &  49.7 & \textbf{79.1} & 36.3 & 20.0 & 27.4 & 17.5&38.3\\
    MVSS\cite{MVSS_2022TPAMI}   &  41.9 & 57.3 & 34.5 & 21.2 & 30.9 & 17.4 &33.9\\
    PSCC\textsuperscript{*}\cite{liu2022pscc}   &  44.5 & 62.5 & 19.4 & 19.8 & 31.0 & 6.0&30.5\\
    Ours   &  \textbf{60.9} & 76.6 & \textbf{38.6} & \textbf{24.2} & \textbf{32.4} & \textbf{23.12}&\textbf{42.6}\\
    \bottomrule
  \end{tabular}
  \end{subtable}%
  \hfill
  \begin{subtable}[t]{0.45\textwidth}
    \centering
    \caption{AUC Score (\%)}
    \begin{tabular}{lccccccc}
    \toprule
    & CASIAV1 & Columbia & Coverage & NIST16 &IMD&CIMD& Average \\
    \midrule
    ViT-S  &78.6 & 68.3 & 62.8 & 70.1 &74.1 &75.7&71.6\\
    SHViT  &85.5 & \textbf{92.5} & \textbf{85.1} & 74.6 &77.4 &79.7&82.4\\
    MVSS   &81.6 & 84.2 & 83.5 & 70.8 &78.9 &71.3&78.4\\
    PSCC\textsuperscript{*}   &77.3 & 82.8 & 79.0 & 69.6 &\textbf{81.6} &66.0& 76.1\\
    Ours   &\textbf{90.6} & 87.9 & 82.4 & \textbf{76.2} & 80.1& \textbf{81.8}&\textbf{83.2}\\
    \bottomrule
  \end{tabular}
  \end{subtable}

  \vspace{1em}

  % Row 2: IoU and Accuracy
  \begin{subtable}[t]{0.45\textwidth}
    \centering
    \caption{IoU Score  (\%)}
    \begin{tabular}{lccccccc}
    \toprule
    & CASIAV1 & Columbia & Coverage & NIST16 &IMD&CIMD& Average \\
    \midrule
    ViT-S  & 26.3 & 15.8 &  2.5 & 10.7 & 13.8&6.7&12.6\\
    SHViT  & 45.1 & \textbf{69.4} & 31.1 & 15.9 &21.8&13.4&32.8  \\
    MVSS   & 37.0 & 46.5 & 26.5 & 16.2 &24.0&13.5&27.3 \\
    PSCC\textsuperscript{*}   & 37.2 & 50.7 & 13.3 & 14.5 &23.2&3.8&23.8 \\
    Ours   & \textbf{56.6} & 69.0 & \textbf{36.2} & \textbf{20.1} &\textbf{27.1}&\textbf{19.0}&\textbf{38.0}  \\
    \bottomrule
  \end{tabular}
  \end{subtable}%
  \hfill
  \begin{subtable}[t]{0.45\textwidth}
    \centering
    \caption{Accuracy Score (\%)}
    \begin{tabular}{lccccccc}
    \toprule
     & CASIAV1 & Columbia & COVERAGE & NIST16 &IMD&CIMD& Average \\
    \midrule
    ViT-S  & 91.2 & 74.4 & 87.2 & 89.7 & 91.3&97.5& 88.5\\
    SHViT  & 93.6 & \textbf{89.9} & 91.6 & 89.7 & 92.8&\textbf{98.1}&92.6\\
    MVSS   & 94.4 & 83.2 & 90.1 & 90.8 & 92.5&96.4&91.2\\
    PSCC\textsuperscript{*}   & 79.4 & 76.1 & 86.6 & 79.1 & 89.1&75.4&80.9 \\
    Ours   & \textbf{95.8} & 86.9 & \textbf{91.9} & \textbf{92.6} & \textbf{93.0}&97.9& \textbf{93.0}\\
    \bottomrule
  \end{tabular}
  \end{subtable}
\label{table_CASIAV2}

\vspace{-0.7em}
\end{table*}

\noindent\textbf{Global Branch} ($x_g$): A fixed number of channels ($C_v$) is allocated to the global branch, which captures long-range dependencies via the recurrent-style attention mechanism of Vision-RWKV6~\cite{duan2024vrwkv}, enabling efficient, context-aware representation learning without explicit pairwise attention. 

\noindent\textbf{Local Branch} ($x_l$): A second group of channels ($C_{con}$) is passed through a depthwise separable convolutional block to capture fine-grained, local spatial patterns. The kernel size(K) of the block varies with the stage number.

\noindent\textbf{Identity Branch} ($x_i$): The remaining channels bypass both global and local processing, preserving raw input features. This facilitates gradient flow through shortcut connections and reduces channel redundancy in high-dimensional space.

\noindent\textbf{Macro Backbone Design.} This is shown in Fig.~\ref{graph_macro}. For an input image \( X \in \mathbb{R}^{H \times W \times 3} \), it is embedded into a (\(\frac{H}{16}, \frac{W}{16}, C_1\)) feature map, fed into the first stage of EfficientRWKV backbone. Each stage of the backbone includes progressive down-sampling through Patch Merging and residual Feed-Forward Network (FFN) blocks to maintain rich multi-scale spatial features. Feature maps processed in stage 2 and 3 are (\(\frac{H}{32}, \frac{W}{32}, C_2\)) and (\(\frac{H}{64}, \frac{W}{64}, C_3\)) respectively. Due to the decrease from quadratic complexity to linear complexity in the foundational attention architectures, the window-based attention strategy in IML-ViT\cite{ma2023imlvit} is no longer necessary.

The output of the three backbone stages yields three feature maps that progressively reduce spatial resolution.

\subsection{Multi-Scale Decoder with Loss Supervision}

The multi-scale features are then passed to a multi-branch decoder head. The decoder outputs the fused prediction as well as intermediate predictions at each resolution. Using multi-scale features directly from different stages of the backbone offers several advantages over extracting only the final feature and relying on a separate Simple Feature Pyramid Network (as done in IML-ViT\cite{ma2023imlvit}). First, the features at early stages inherently preserve finer spatial details and shallow representations. Second, this direct multi-stage integration enables more efficient gradient flow, which improves learning stability and accelerates convergence. Lastly, it reduces computational overhead by eliminating the need for a separate SFPN.

The total training loss $\mathcal{L}_{\text{total}}$ consists of edge-aware supervision from IML-ViT\cite{ma2023imlvit} and multi-scale prediction loss.
\vspace{-0.5cm}

\begin{eqnarray*}
\begin{aligned}
\mathcal{L}_{\text{total}} &= \sum_{i=1}^{3} \lambda_i \cdot \text{BCE}(\hat{M}_i, M_i) 
+ \lambda_{\text{edge}} \cdot \sum_{i=1}^{3} \text{BCE}(\hat{M}_i, M_i; E_i) \\
&\quad + \lambda \cdot \text{BCE}(\hat{M}_{\text{final}}, M) 
+ \lambda_{\text{edge}} \cdot \text{BCE}(\hat{M}_{\text{final}}, M; E)
\end{aligned}
\end{eqnarray*}

In the above formula, $\hat{M}_i$ denotes the predicted mask at the $i$-th intermediate resolution. $M_i$ is the ground-truth mask downsampled to the same resolution as $\hat{M}_i$. $E_i$ is the corresponding edge mask and $\hat{M}_{\text{final}}$ is the upsampled final prediction. $M$ and $E$ are the ground-truth full-resolution mask and edge map, respectively. $\lambda_i$ are the weights for multi-scale prediction losses. $\lambda_{\text{edge}}$ is the edge loss hyperparameter.

\vspace{-0.3cm}
\section{Experiment}

\subsection{Implementation Details}

\noindent\textbf{EfficientRWKV.} The EfficientRWKV backbone uses a pretrained checkpoint obtained by pre-training for 1,000 epochs on ImageNet-1K under a teacher–student distillation regime, with a TResNet-L network\cite{ridnik2020tresnethighperformancegpudedicated} providing soft targets according to the DeiT distillation protocol\cite{touvron2021trainingdataefficientimagetransformers}. Training is performed with an effective batch size of $64$, using the AdamW\cite{loshchilov2019decoupledweightdecayregularization} optimizer with an initial learning rate of $1\times10^{-4}$ and a cosine decay schedule\cite{loshchilov2017sgdrstochasticgradientdescent}.

\noindent\textbf{ViT-S.}
The vision transformer in IML-ViT\cite{ma2023imlvit} method is modified to be ViT-S, a lighter version of ViT with checkpoint ViT-S/16 from DINO\cite{caron2021emerging}, to ensure fairness in comparison.

\noindent\textbf{SHViT.}
We also replace the attention module in IML-ViT\cite{ma2023imlvit} with SHViT-S4\cite{yun2024shvit}, by Yun and Ro et al. 

\noindent\textbf{MVSS\cite{MVSS_2022TPAMI} \& PSCC\cite{liu2022pscc}.}
We employ IMDL-BenCo\cite{ma2025imdl}, a comprehensive platform for IMDL tasks, to experiment on these two models. PSCC has 512×512 instead of 1024×1024 inputs due to its high demand for GPU memory.

\noindent\textbf{Training/Testing Dataset.}

\noindent1. CASIAV2-training: Aligning with IML-ViT\cite{ma2023imlvit}, we use entire CASIAV2 for training, other datasets for testing.

\noindent2. SIF-training: In the second experiment, SIF dataset is split with a 6:1 ratio for training and testing.

\noindent\textbf{Evaluation Criteria.}
We evaluate all models using four metrics: Pixel-F1 score, Area Under the ROC Curve (AUC), Intersection over Union (IoU), and Pixel-Accuracy.

\subsection{Comparison with SoTAs}
\noindent\textbf{Quantitative Results.} Under the CASIAV2-training protocol (Tab.~\ref{table_CASIAV2}), our model has consistently strong performance across all metrics. For F1 and IoU, it has increased performance of +4.3\% and +5.2\% over the best SOTA model tested. 

\begin{table}[t]
\scriptsize 
\setlength{\tabcolsep}{3pt}
\centering
\captionsetup{justification=centering}
\caption{\textbf{SIF-training at different resolutions}}
\captionsetup{justification=centering}
\vspace{-0.3em}

\begin{subtable}[t]{0.49\linewidth}
\centering
  \caption{Resolution 1024×1024(\%)}
  \label{table_SIF1024}
  \begin{tabular}{lcccc}
    \toprule
         &   F1 &  AUC &  IoU &  ACC \\
    \midrule
    ViT-S  & 23.8 & 84.2 & 16.9 & 95.1 \\
    SHViT  & 55.9 & 95.4 & 43.9 & 96.9 \\
    MVSS   & 61.5 & 92.7 & 48.5 & 96.9 \\
    PSCC   & 54.5 & 93.4 & 41.5 & 95.2 \\
    Ours   & \textbf{68.1} & \textbf{97.3} & \textbf{56.2} & \textbf{97.4} \\
    \bottomrule
  \end{tabular}
\end{subtable}
\hfill
\begin{subtable}[t]{0.49\linewidth}
\centering
  \caption{Resolution 2048×2048(\%) }
  \label{table_SIF2048}
  \begin{tabular}{lcccc}
    \toprule
     &   F1 &  AUC &  IoU &  ACC \\
    \midrule
    ViT-S  & 39.9 & 91.9 & 29.2 & 96.3 \\
    SHViT  & 59.0 & 94.6 & 47.6 & 97.1 \\
    MVSS   & 65.0 & 92.9 & 51.3 & 97.1 \\
    PSCC   & - & - & - & - \\
    Ours   & \textbf{78.0} & \textbf{98.8} & \textbf{67.3} & \textbf{98.2} \\
    \bottomrule
  \end{tabular}
\end{subtable}

\vspace{-1em}

\end{table}

Tab.~\ref{table_SIF2048} presents the evaluation results on the SIF dataset trained and tested at resolution of 2048×2048. Our proposed model again demonstrates clear superiority across all four metrics, with a +13\% difference compared to the best model. This increase confirms the effectiveness of our architecture in capturing fine-grained manipulations in large images.

\begin{figure}[htb]
  \centering
  \includegraphics[width=\linewidth]{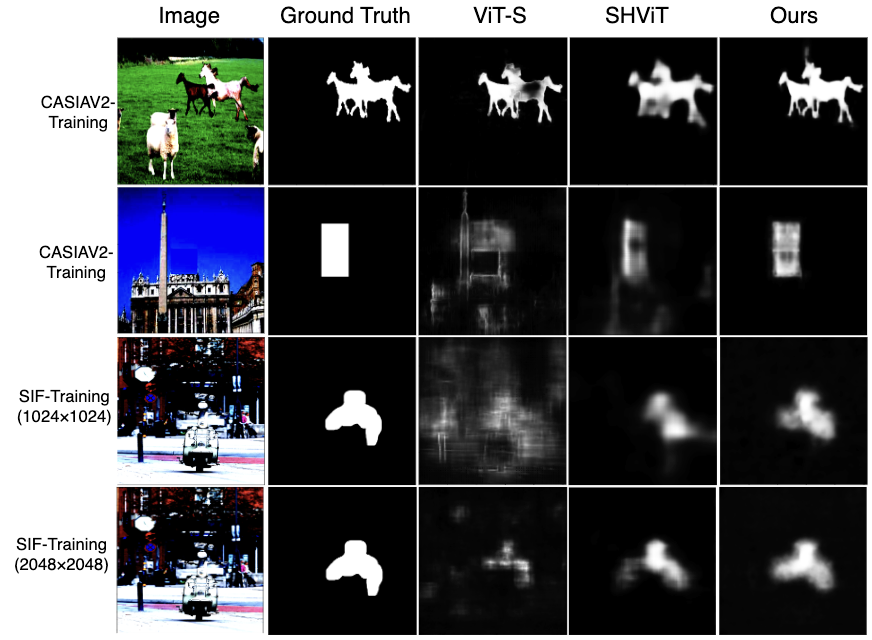}
  \caption{\textbf{Qualitative Comparison}}
  \label{visual_sota}
\end{figure}

\noindent\textbf{Qualitative Results.} The qualitative comparison shown in Fig.~\ref{visual_sota} highlights the validity of our proposed model across both CASIAV2-trained and SIF-trained settings. In the first two rows (CASIAV2-training), our method shows significantly more accurate localization than ViT-S and sharper predictions than SHViT\cite{yun2024shvit}. In the bottom row with higher resolution, our model leverages the additional details, becoming noticeably better aligned with ground truth.

\begin{table}[tbh]
  \centering
  \scriptsize
  \captionsetup{justification=centering}
    \caption{\textbf{Model Complexity} (*: Checkpointing used)}
  \setlength{\tabcolsep}{4pt}
  \label{table_complexity}
  \begin{tabular}{lccccc}
    \toprule
    & & \multicolumn{2}{c}{1024$\times$1024} & \multicolumn{2}{c}{2048$\times$2048} \\
    \cmidrule(lr){3-4} \cmidrule(lr){5-6}
            & Params(M) & GFLOPS & Speed(img/s) & GFLOPS & Speed(img/s) \\
    \midrule
    ViT-S  &        22.6 &  128.2 &   37.8 &  512.7\textsuperscript{*} &    9.3\textsuperscript{*} \\
    SHViT  &        19.5 &   95.6 &   \textbf{71.7} &  382.4 &   16.9 \\
    MVSS   &       142.8 &  654.3 &   11.0 & 2617.8 &    2.4 \\
    PSCC  &         3.7 &  116.2 &   13.8 &     --  &    --  \\
     Ours   &        19.8 &   \textbf{21.7} &   58.5 &   \textbf{86.7} &   \textbf{23.0} \\
    \bottomrule
  \end{tabular}
\end{table}

\noindent\textbf{Efficiency.} As shown in Tab.~\ref{table_complexity}, our method achieves a superior computation–throughput trade-off among state-of-the-art approaches. At 1024×1024 resolution, it requires only approximately \textbf{22.7\%} of the computational resources as SHViT\cite{yun2024shvit} (lightest model tested) used and around \textbf{3.3\%} as the SOTA model MVSS\cite{MVSS_2022TPAMI} used. With ultra-high resolution at around 2048×2048, our model incurs just $86.7$ GFLOPs, still maintaining the usage of around \textbf{22.7\%} of the computational resources as the lightest comparison model.

\begin{table}[h]
  \centering
  \setlength{\tabcolsep}{3pt}
  \footnotesize
  \caption{\textbf{Model Ablation Studies} \\ *: See Appendix Table \ref{table_weight}}
  \begin{tabular}{lcccc}
    \toprule
      & \textbf{F1} & \textbf{AUC} & \textbf{IoU} & \textbf{ACC} \\
    \midrule
    Loss Weights Set 1 \textsuperscript{*}                       & 65.7 & 97.1 & 54.1 & 97.3 \\
    Loss Weights Set 2 \textsuperscript{*}                        & 65.2 & 96.9 & 53.6 & 97.2 \\
    Loss Weights Set 3 \textsuperscript{*}                 & 65.8 & 97.0 & 54.6 & \textbf{97.4} \\
    Pretrained w/o Knowledge Distillation         & 63.8 & 96.5 & 52.9 & \textbf{97.4} \\
    Freeze Global Branch Weights              & 57.3 & 96.8 & 46.3 & 96.8 \\
    Freeze Local Branch Weights               & 64.2 & 97.0 & 53.0 & 97.2 \\
    Final Setup                                 & \textbf{68.1} & \textbf{97.3} & \textbf{56.2} & \textbf{97.4} \\
    \bottomrule
  \end{tabular}
  
  \label{table_abl}
\end{table}

\vspace{-0.3cm}
\subsection{Ablation}
\noindent\textbf{Multi-Loss Weights.}
We tested other sets of loss weight and our chosen set achieves the best result (Tab. \ref{table_abl}).

\noindent\textbf{Pre-training.}
Two pretrained checkpoints are tested: One is pretrained with knowledge distillation and the other without. Employing this special pre-training technique boosts our model performance by 4.3\% in F1.

\noindent\textbf{Ablation on EfficientRWKV.}
According to Tab.~\ref{table_incre} row 1 and 2, EfficientRWKV is proven to be a more effective replacement of ViT-S. We also conducted two ablation trials to validate the effectiveness of the branches in EfficientRWKV.

\begin{table}[h]
  \centering
  \footnotesize
    \caption{\textbf{Model Incremental Ablation Studies}}
  \label{table_incre}
  \begin{tabular}{lcccccc}
    \toprule
     Backbone & SFPN & Multi-S Loss & \textbf{F1} & \textbf{AUC} & \textbf{IoU} & \textbf{ACC} \\
    \midrule
    ViT-S                 & Yes & No  & 23.8 & 84.2 & 16.9 & 95.1 \\
    Effi-RWKV                 & Yes & No  & 57.6 & 97.1 & 47.0 & 97.1 \\
    Effi-RWKV                 & No  & No  & 61.6 & 95.5 & 50.0 & 97.0 \\
    Ours      & No  & Yes & \textbf{68.1} & \textbf{97.2} & \textbf{56.2} & \textbf{97.4} \\
    \bottomrule
  \end{tabular}
  \vspace{0em}
\end{table}

\vspace{-0.5cm}
\section{Conclusion}
In this paper, we introduce SIF, the first high–resolution semantic inpainting forgery dataset and EfficientIML with a lightweight three-stage backbone of linear complexity and a multi-scale decoder under hierarchical loss supervision. Extensive experiments on SIF and standard benchmarks demonstrate that our method achieves state-of-the-art localization performance and runtime efficiency.

\appendix
\twocolumn[
\begin{center}
    \LARGE \textbf{Appendix}
\end{center}
\vspace{1em}]
\section{SIF Dataset Related}

\begin{figure}[h]
\centering
\includegraphics[width=\linewidth]{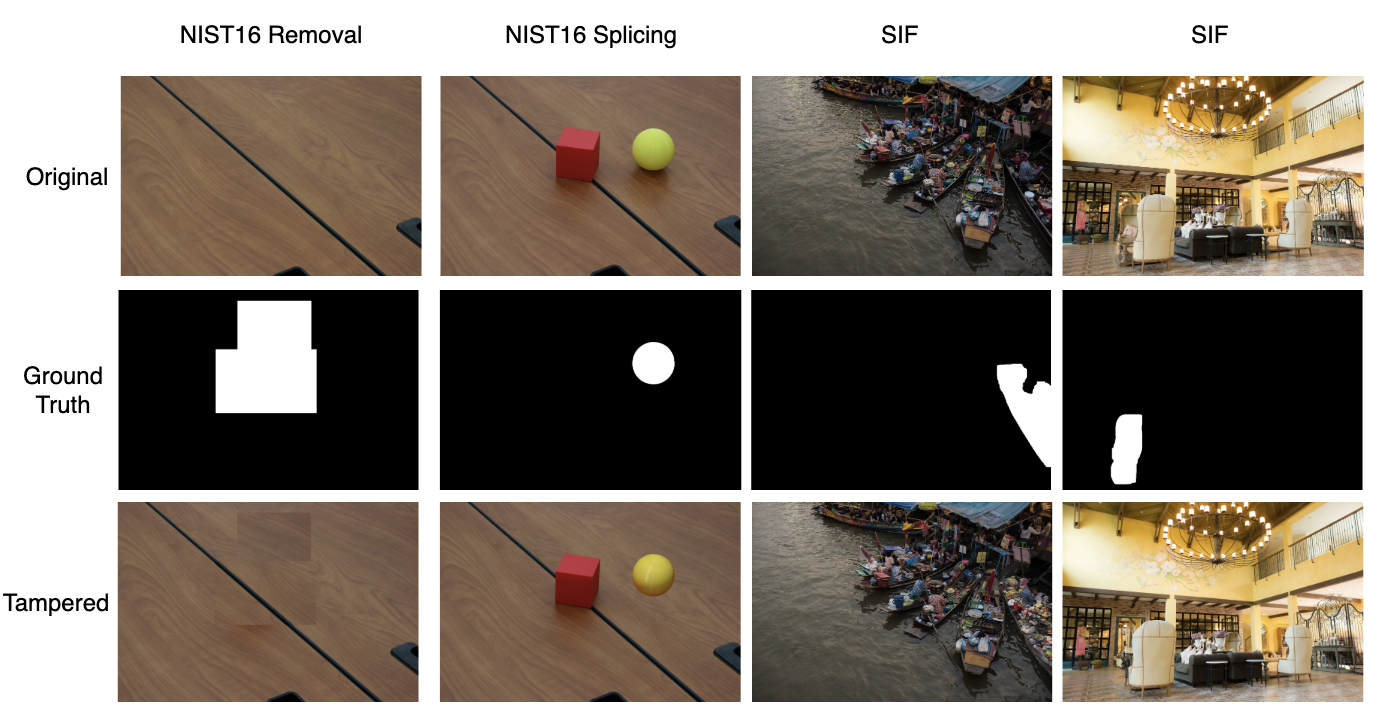}
\caption{\textbf{Comparison between SIF and NIST16}}
\label{graph_dataset_compare}
\end{figure} 

Fig.~\ref{graph_dataset_compare} highlights a qualitative difference between NIST16 (an existing and widely-used high-resolution dataset) and our SIF dataset. In NIST16 (left two columns), manipulated regions are restricted to simple geometric primitives that sometimes bear little relation to image content. By contrast, SIF's context-aware masks trace the contours of actual scene elements, so the inpainted areas correspond to semantically meaningful objects (e.g. furniture or vessels). This context-aware masking is far more closely aligned with real-world scenarios, producing highly realistic forgeries and yielding a far more effective and valuable training dataset for high-resolution localization methods.

\begin{table}[h]
\footnotesize
\centering
\caption{\textbf{Current IML Datasets / SIF Dataset}}
\vspace{1em}
\begin{tabular}{ccc}
\toprule
\textbf{Dataset} & \textbf{\# Manipulated Images} & \textbf{Resolution} \\
\midrule
CASIA V1\cite{Dong2013}         & 921                             & $\sim$384×256    \\
CASIA V2\cite{pham2019hybrid}         & 5,123                           & 384×256 to 800×600   \\
Columbia\cite{hsu06crfcheck}         & 180                             & 757×568 to 1152×768  \\
NIST16\cite{518026}           & 564                             & Up to 3744×5616 \\
DEFACTO\cite{DEFACTODataset}          & 229,000                         & $\sim$640×480 \\
COVERAGE\cite{wen2016}         & 100                             & 235×158 to 752×563     \\
ForenSynths\cite{wang2019cnngenerated}      & 100,000                         & 256×256              \\
CSI-IMD\cite{10943879}  & 1000                     & 400×296 to 600×800    \\
\midrule
SIF             & 1,228                         & $\sim$1800×1200        \\
\bottomrule
\end{tabular}
\vspace{1em}
\label{tab:dataset_comparison}
\end{table}

\begin{figure}[h]
\centering
\includegraphics[width=0.7\linewidth]{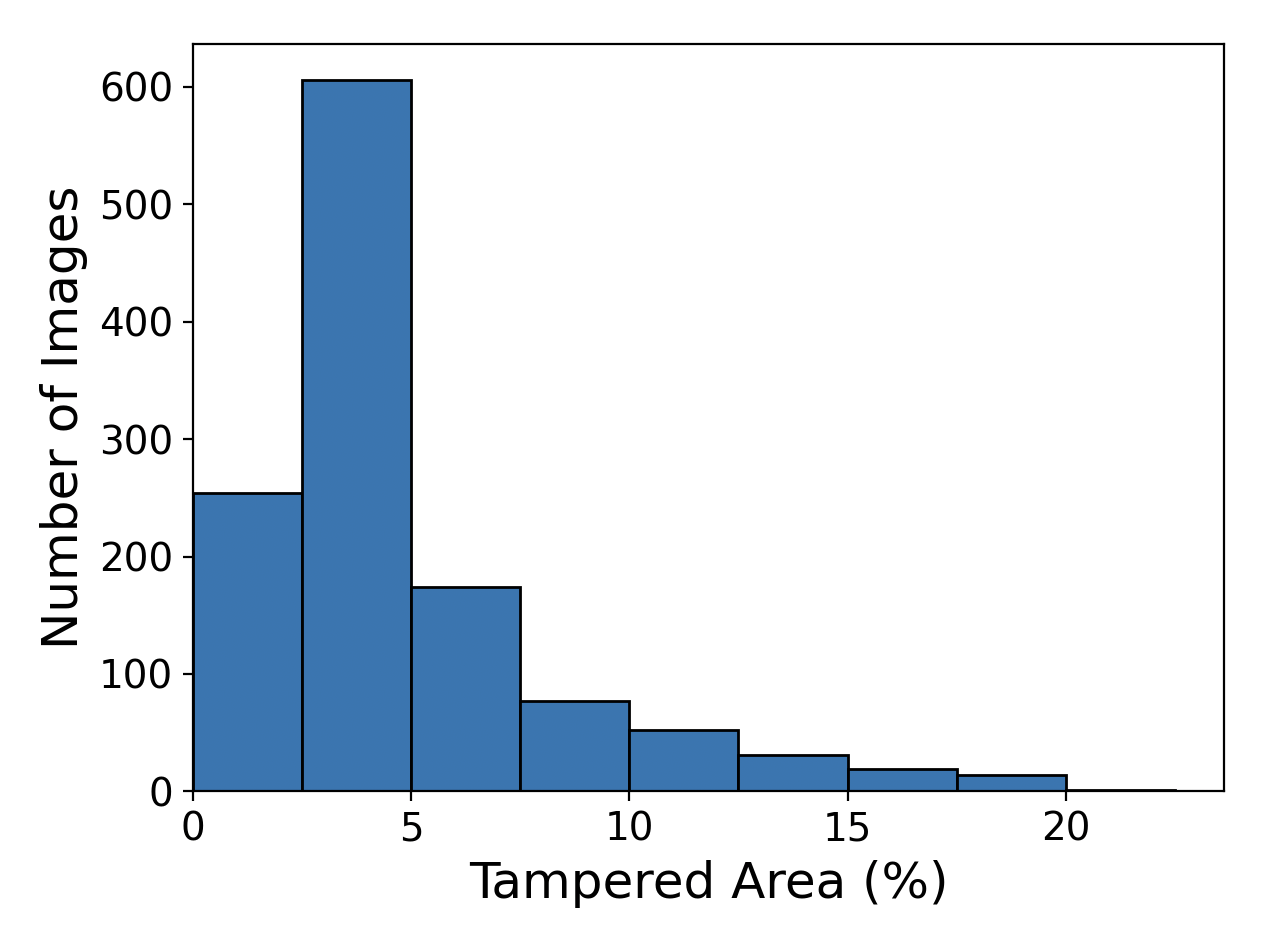}
\caption{\textbf{Tampered Region Ratio of SIF}}
\label{graph_tpRaio}
\end{figure}

\section{EfficientRWKV Related}

\subsection{Additional Details about Model}
The three channel sizes used in each stage are [$C_1= 200$, $C_2= 376$, $C_3= 448$].

\begin{table}[h]
\centering
\caption{\textbf{Channel Allocation per Stage}}
\begin{tabular}{cccc}
\toprule
\textbf{Stage} & $C_v$ & $C_{\text{con}}$ & $C_i$ \\
\midrule
1 & $0.8 \times C$ & $0.2 \times C$ & $0 \times C$ \\
2 & $0.7 \times C$ & $0.2 \times C$ & $0.1 \times C$ \\
3 & $0.6 \times C$ & $0.3 \times C$ & $0.1 \times C$ \\
\bottomrule
\end{tabular}
\end{table}

\subsection{Additional Details about Ablation Experiemnets}

Tab. \ref{table_weight} contains the details of the three loss weight sets used in ablation studies. Our chosen weighting scheme achieves the best performance, delivering an approximately 2.5\% improvement in F1 score over other configurations. These findings indicate that more empha- sis should be given to the deeper layers, although taking multi-scale loss benefits the performance of the model. 
\begin{table}[h]
\footnotesize
  \centering
    \caption{\textbf{Weights for Model Outputs in Ablation Studies}}
  \label{table_weight}
  \begin{tabular}{lcccc}
    \toprule
     Set Number  & Layer 1 & Layer 2 & Layer 3 & Fused Output \\
    \midrule
    Loss Weights Set 1    &  0.25 & 0.35 & 0.45 & 1.0 \\
    Loss Weights Set 2                 &  0.35 & 0.35 & 0.35 & 1.0 \\
    Loss Weights Set 3                 &  0.5 & 0.5 & 0.5 & 0.5    \\
    Final Setup                  &  0.15 & 0.35 & 0.55 & 1.0 \\
    \bottomrule
  \end{tabular}
\end{table}

\begin{figure}[h]
\centering
\includegraphics[width=\linewidth]{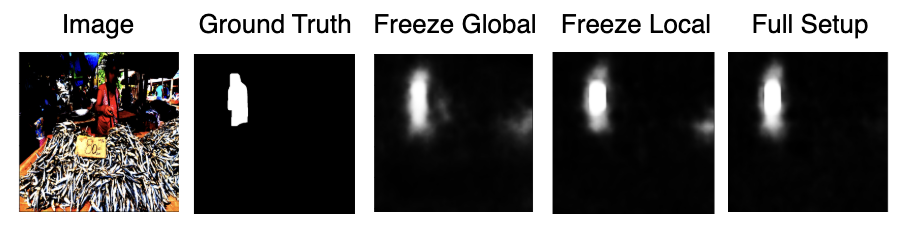}
\caption{\textbf{Full Set up and freezing Local/Global Branches}}
\label{fig_abl}
\end{figure}

For local/gloabl branch studies, we also include a qualitative result to show the difference in Fig.~\ref{fig_abl}, corresponding to the decreased value in testing metrics. When the global branch is frozen, the predicted mask lacks confidence and exhibits excessively blurred boundaries; moreover, false positives appear to the right of the whole prediction. Conversely, freezing the local branch yields a mask with a highly confident predicted area, also accompanied by false positives towards the right. Only when both branches operate together does the model reconcile coarse, context‐driven localization with fine, edge‐aware refinement, eliminating the majority of false positives and producing a sharply delineated mask that faithfully matches the ground truth.

\section{RWKV Basics}
Our global attention branch is based on RWKV. Specifically, the Bi-WKV module computes global attention as:
\begin{eqnarray*}\label{eq:rwkv-wkv}
\mathrm{wkv}_t
=\frac{\sum_{i=0,i\neq t}^{T-1}\exp\bigl(-\tfrac{|t-i|-1}{T}w + k_i\bigr)\,v_i
\;+\;\exp(u + k_t)\,v_t}
{\sum_{i=0,i\neq t}^{T-1}\exp\bigl(-\tfrac{|t-i|-1}{T}w + k_i\bigr)
\;+\;\exp(u + k_t)}
\end{eqnarray*}

WKV recurrence\cite{duan2024vrwkv} is used here in Equation~\eqref{eq:rwkv-wkv}. $w, u \in \mathbb{R}^{C_v}$ are learnable channel-wise decay and bias vectors, and $k_i, v_i$ are the key/value features at position $i$, yielding $\mathcal{O}(T C_v)$ complexity.

\section{Related Work}

\subsection{Image Forgery Dataset}
\label{rw_dataset}
Conventional image forgery techniques primarily involve copy-move, splicing, and object removal. Accordingly, many image forgery detection datasets are constructed following these manipulation types, including CASIA V1/V2\cite{Dong2013}\cite{pham2019hybrid}, Columbia\cite{hsu06crfcheck}, NIST16\cite{518026}, DEFACTO\cite{DEFACTODataset}, and Coverage\cite{wen2016}. Recent deep learning-generated datasets such as ForenSynths\cite{wang2019cnngenerated} have trivial and unified artifacts that are often easier to detect, as their creator notices. On the other hand, our proposed dataset employs Stable Diffusion\cite{rombach2021highresolution} for inpainting, attempting to model the challenge of detecting more subtle AI-generated traces.
Moreover, most existing image forgery datasets consist predominantly of low/medium-resolution images, which may not adequately represent the quality of manipulations generated by modern generative models and advanced editing tools. For instance, the CASIA v2.0 dataset contains images with average resolutions below 500×500, and the Columbia dataset typically features images below 1000×1000 pixels. In this work, we introduce a dataset comprising images with resolutions around 1800×1200, thereby providing a more suitable benchmark for evaluating image forgery localization models under high-resolution settings.

\subsection{CNN Based Image Forgery Localization} Before the rise of Vision Transformers, Convolutional Neural Networks (CNNs) were the dominant architecture for the task of image manipulation localization. Early approaches like ManTra-Net\cite{8953774} focused on learning local artifact features through an encoder-decoder framework that combines manipulation traces and classification clues. PSCC-Net\cite{liu2022pscc} advanced this by introducing a pyramid structure and spatial-channel correlation modules to capture multi-scale inconsistencies. MVSS-Net\cite{MVSS_2022TPAMI} has further improved performance by integrating both RGB content and noise streams, guided by edge supervision, to highlight manipulation boundaries. These CNN-based methods laid the groundwork for spatially aware manipulation localization, but often struggled with capturing relatively long-range dependencies and generalizing across diverse manipulation types.

\subsection{ViT Based Image Forgery Localization}
Significant advancements in image manipulation localization are made possible by leveraging the power of Vision Transformer. Omni-IML\cite{qu2024omniimlunifiedimagemanipulation} presents a unified framework capable of localizing a wide range of manipulation types. However, the use of large-scale ViT backbones leads to substantial computational costs. Similarly, IML-ViT\cite{ma2023imlvit} (which serves as our baseline), along with other ViT-based models, faces scalability challenges, particularly when processing high-resolution images, due to their high computational complexity and memory consumption. With the demand for processing images with higher resolutions, significant efforts are made to reduce model complexity, such as token pruning presented by Evo-ViT\cite{evo-vit}, or window-based attention layers in IML-ViT\cite{ma2023imlvit}. However, these techniques tend to have limited effects as opposed to strategies that directly modify the model's backbone architecture.

%\vfill
%\pagebreak

\vfill\pagebreak

\begingroup
\footnotesize  
\bibliographystyle{IEEEbib}
\bibliography{strings,refs}
\endgroup

\end{document}